\documentclass[letterpaper, 10 pt, conference]{icra17}  

\IEEEoverridecommandlockouts                              

\usepackage{graphicx}	
\usepackage{wrapfig}
\usepackage{multirow}
\usepackage{amsmath}
\usepackage[ruled,vlined]{algorithm2e}
\usepackage{color}

\usepackage{graphics} 
\usepackage{overpic}
\usepackage{epsfig} 
\usepackage{mathptmx} 
\usepackage{times} 
\usepackage{amsmath} 
\usepackage{amssymb}  

\usepackage{algorithmicx}
\usepackage{algpseudocode}
\newtheorem{theorem}{Theorem}[section]

\newtheorem{defn}[theorem]{Definition}
\newtheorem{property}[theorem]{Property}

\newcommand{\qed}{\nobreak \ifvmode \relax \else
      \ifdim\lastskip<1.5em \hskip-\lastskip
      \hskip1.5em plus0em minus0.5em \fi \nobreak
      \vrule height0.75em width0.5em depth0.25em\fi}
\usepackage{paralist}
\let\itemize\compactitem
\let\enditemize\endcompactitem
\let\enumerate\compactenum
\let\endenumerate\endcompactenum

\title{Caging Loops in Shape Embedding Space: Theory and Computation}

\author{Jian Liu$^{1}$, Shiqing Xin$^{1}$, Zengfu Gao$^{1}$, Kai Xu$^{2}$, Changhe Tu$^{1}$ and Baoquan Chen$^{1}$
\thanks{$^{1}$Jian Liu, Shiqing Xin, Zengfu Gao, Changhe Tu and Baoquan Chen are with School of Computer,
        Shandong University, 266237 Qingdao, P. R. China.
        $^{2}$Kai Xu is with School of Computer,
        National University of Defense Technology, 410073 Changsha, P. R. China.
       This work was supported by National 973 Program (2015CB352501), and NSFC (61332015, 61772318, 61572507, 61532003, 61622212, 61772016).}
}
\begin{document}

\maketitle

\begin{abstract}
We propose to synthesize feasible caging grasps for a target object
through computing \emph{Caging Loops}, a closed curve defined in the \emph{shape embedding space}
of the object.
Different from the traditional methods,
our approach \emph{decouples} caging loops from
the surface geometry of target objects through working in the embedding space.
This enables us to synthesize caging loops encompassing multiple topological holes,
instead of always tied with one specific handle
which could be too small to be graspable by the robot gripper.
Our method extracts caging loops through a topological analysis of the distance field
defined for the target surface in the embedding space, based on a rigorous theoretical study
on the relation between caging loops and the field topology.
Due to the decoupling, our method can tolerate incomplete and noisy surface geometry of
an unknown target object captured on-the-fly.
We implemented our method with a robotic gripper and demonstrate through extensive experiments
that our method can synthesize reliable grasps for objects with complex surface geometry and topology
and in various scales.

\end{abstract}

\section{Introduction}\label{sec:intro}
As an important type of robot grasping, caging grasps~\cite{Rodriguez-2011FromCT,Wan-2013AN,Diankov-2008ManipulationPW},
as compared to force-closure grasps~\cite{Zhu-PlanningFG2004,Ding-Computing3O2000,Borst-GraspingTD2003,Ferrari-PlanningOG1992},
are advantageous in handling target objects with unknown or uncertain surface geometry and/or friction properties.
This makes caging grasps more practically applicable in a wide spectrum of real scenarios.
We are especially interested in a simple yet effective type of caging grasp formed by \emph{caging loops}.
A caging loop is a closed curve in three dimensional space computed around some part of the target object and used to guide robot grippers to form a caging grasp.

\begin{figure}[t]
	\centering	
	\includegraphics[width=0.95\columnwidth]{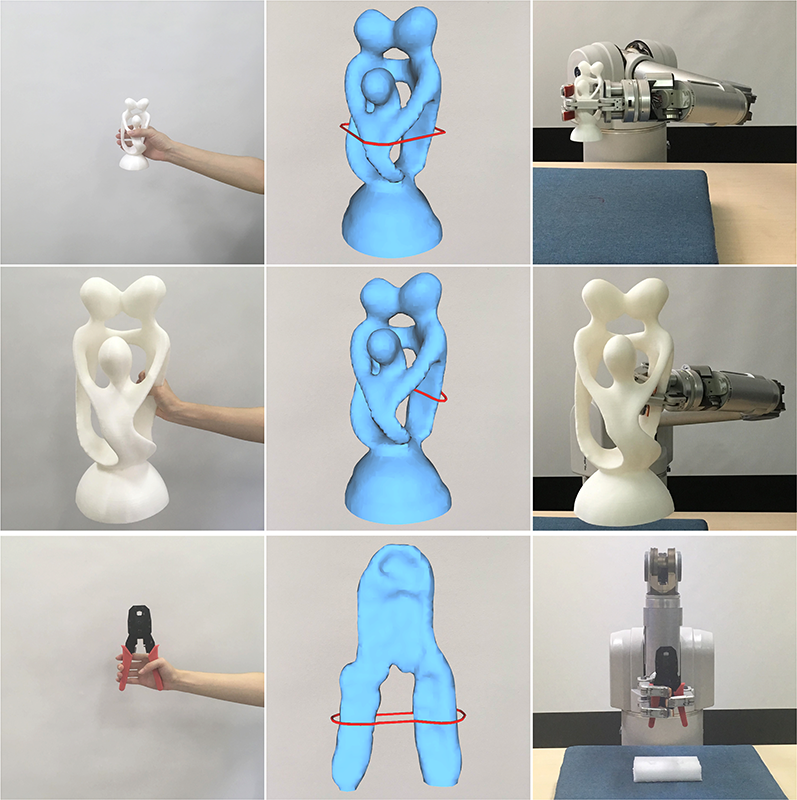}
	\caption{Grasping a 3D-printed Indonesian-Lady model (the top and middle row) in two sizes.
            Our method is able to synthesize caging loops (red circles) encompassing multiple topological handles,
            when the object is too small to be grasped on one handle (top row).
            When the object is large, our method naturally grasps one handle (middle row).
            The two cases are integrated seamlessly in method.
            The bottom row shows how a caging loop computed in the embedding encloses
            the two handles of a pliers. The 3D objects are acquired by two RGBD cameras
            and reconstructed on-the-fly (middle column).
            As a reference, a human grasp is shown to the left for each object.}
\label{fig:teaser}
\vspace{-2mm}
\end{figure}

\begin{figure}[t]
	\centering	
	\includegraphics[width=0.95\linewidth]{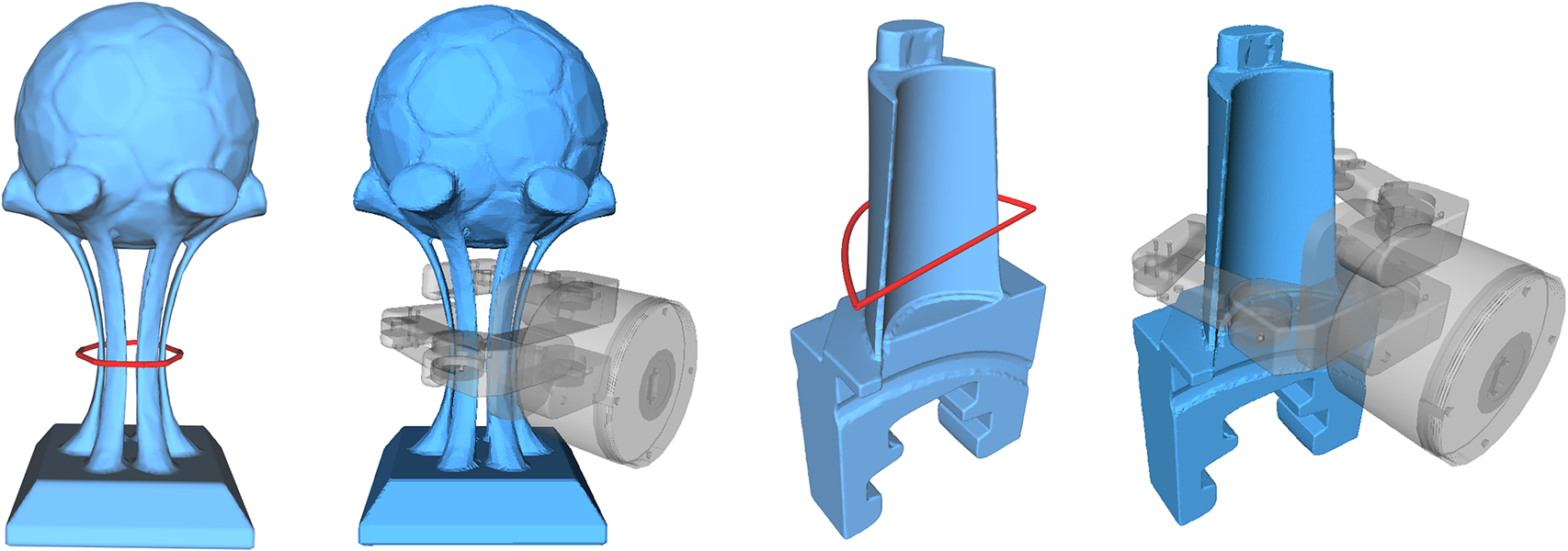}
    \parbox[t]{1.0\linewidth}{\relax
    \hspace{19 mm} (a) \hspace{30 mm} (b)}\\
    \parbox[t]{1.0\linewidth}{\relax }\\
    \includegraphics[width=0.95\linewidth]{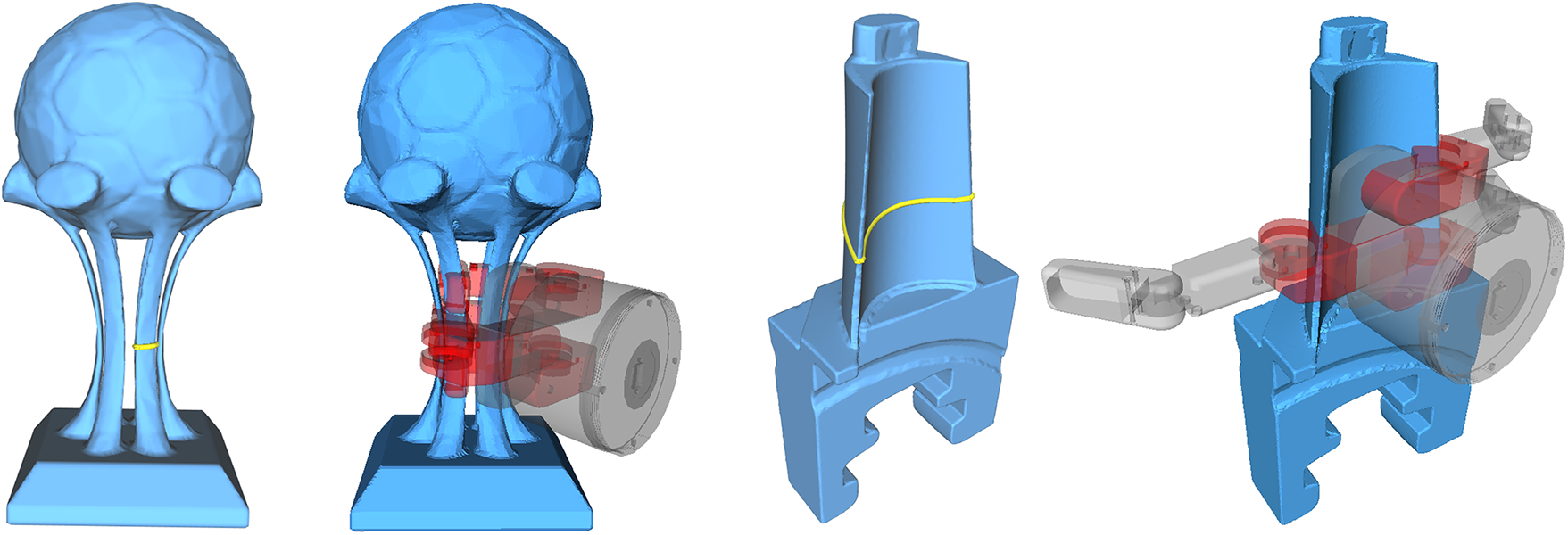}
    \parbox[t]{1.0\linewidth}{\relax
    \hspace{19 mm} (c) \hspace{30 mm} (d)}\\
    \caption{By decoupling caging loops from target surfaces, our method
            synthesizes feasible caging grasps for objects containing tiny topological handles (a) or presenting
            concave surface geometry (b); see the red circles and the corresponding grasps to the right.
            In contrast, the loops (yellow circles in c and d)
            computed over the target surfaces incur gripper-object collision; see the gripper parts in red color in the bottom row.}
    \label{fig:topologyAndGeometry}
\vspace{-2mm}
\label{fig:scale}
\end{figure}

Existing methods on 3D caging grasp 
are based either on the geometric (e.g.~\cite{Zarubin-2013CagingCO})
or the topological (e.g.~\cite{Pokorny-2013GraspingOW}) information of the target surface,
or even both~\cite{Kwok-2016RopeCA}.
A common issue to these methods is that the computed caging curves seriously depend on topological and geometrical features of objects,
while being oblivious to the relative size between the target object and the gripper.
Taking the genus-4 Indonesian-Lady model
in Fig.~\ref{fig:teaser} for example.
The six handles on the model are all seemingly good candidates for grasping.
However, when the size of the model is too small compared to the robot gripper,
these handles will no longer be graspable since the holes may be too small for the fingers
to pass through.
In such case, a more feasible grasp would be enclosing the object with a loop encompassing
multiple handles (see Fig.~\ref{fig:teaser}(top) and Fig.~\ref{fig:scale}(a)).

Another issue with geometry-based caging curves is that they easily lead to non-convex spatial curves
which are not suited for guiding the gripper configuration.
The example in Fig.~\ref{fig:scale}(d) demonstrates such case, where
the gripper penetrates into the object due to the non-convexity of the caging loop.
Estimating a convex hull for the spatial loop still cannot guarantee a
penetration-free configuration.

\subsection{Motivation And Contribution}


These examples motivated us in seeking to
``fill up'' those small topological holes and ``smooth out'' the geometric details
on the target surface, before computing caging curves.
Therefore, we advocate computing caging loops
in the \emph{embedding space} of the target surface,
through a topological analysis of the distance field defined for the target surface in the embedding space.

We conduct a theoretical study on the fundamental relationship
between caging loops and Morse singularities (including minimal, maximal and saddle points)
of a spatial distance function.
Based on that, we develop an algorithm of caging loop extraction through
saddle point detection and analysis, within a proper grasping space defined in account of the gripper size.
Working with a distance function defined in the embedding space
naturally decouples the shape of caging loops from
the geometric details of the target surface, while still keeping them aware of the overall shape
of the target object.
The caging loops are properly placed and scaled
based on the relative size of the gripper against the target shape,
rather than always tied with a specific handle as in traditional approaches.

Another benefit of working in embedding space is the tolerance of incomplete and noisy
surface geometry of the target object. This makes our method especially suited for
synthesizing grasps for unknown objects which are captured and reconstructed on-the-fly,
with a minimal effort of robot observation.
In our implementation (see Fig.~\ref{fig:overview}), two depth cameras are deployed
to capture the target object from two (front and back) views.
Even with such a sparse capturing and low-quality reconstruction, our method can still synthesize
feasible caging loops for robust grasping.


We found this simple idea leads to a robust and efficient algorithm,
with theoretical guarantees.
We implemented our algorithm in a grasping system composed of a Barrett WAM robotic arm
with a three-finger gripper and two Xtion Pro RGB-D cameras.
Only depth images are used for reconstructing the target surface
based on the depth fusion technique~\cite{Newcombe-2011KinectFusionRD}.
We have conducted numerous evaluations
with both synthetic and real examples to evaluate the performance of our method.
We show that our system is able to robustly grasp objects with complex surface geometry and topology
and in various scales.

Our work makes the following contributions:
\begin{itemize}
\item We propose a novel method for caging grasp synthesis through topological analysis of shape-aware distance field defined in shape embedding space. The method is able to generate relative-scale-aware caging loops for unknown objects captured on-the-fly.
\item We provide a rigorous study on the relation between the topology of distance field and caging loops,
    based on Morse theory, and derive a robust algorithm for caging loop estimation. We also provide a handful of provably effective techniques to reduce the computational cost of our method.
\item We implement our method in a grasping system using robot gripper, and conduct thorough evaluations and comparisons with both synthetic and real objects.
\end{itemize}

\subsection{Related Work}
Robot grasping is a long-standing yet actively studied research topic in the fields of robotics and vision.
Force-closure and caging are two typical approaches that have been developed to synthesize grasps.
Force-closure methods~\cite{Bicchi-OnTC1995,Miller-GraspitAV2004,Liu-QualitativeTA1999,Howard-OnTS1996,Ding-Computing3O2000,Zhu-PlanningFG2004} concentrate on finding a stable grasping configuration for the grippers
where a mechanical equilibrium is achieved.
The advantage of such approach is that the synthesized grasps are usually physically feasible.
The method, however, requires that the 3D shape of the target model is known \textit{a priori} and cannot tolerate
much the surface defect such as missing data.
Furthermore, the contact area between the gripper and the target surface is often small, leading to unsteady grasps.

Caging grasps~\cite{Diankov-2008ManipulationPW}, as compared to force-closure ones,
seeks for a sufficiently large contact area and thus are deemed to have better stability,
although they are not designed to directly reach a mechanical equilibrium.
The key benefit of caging~\cite{Diankov-2008ManipulationPW} is that
it is robust to surface uncertainty and imperfection.
This makes it especially applicable to unknown objects
being captured and reconstructed on-the-fly.
Some works studied the computation of planar cages in 2D space for planar
objects~\cite{Rimon-1999CagingPB,Pipattanasomporn-2006TwofingerCO,Pipattanasomporn-2011TwoFingerCO,Vahedi-2008CagingCP}.
Most existing approaches to 3D object caging rely on the topological structure of the target surface~\cite{Pokorny-2013GraspingOW,Dey-2010ApproximatingLI,Stork-2013IntegratedMA}.
Some further take geometry information into account~\cite{Kwok-2016RopeCA}.
However, such approaches cannot compute a caging loop encompassing
multiple handles or deal with different relative scales between the gripper and the target object.

The Morse theory, as a connection between geometry and topology, has been widely utilized
in the graphics and visualization fields~\cite{Bremer-2004ATH,Ni-2004FairMF}.
In our approach, the core algorithmic step is to find a caging loop
according to a $p$-based distance field, where $p$ is a point in the grasping space.
At this point, the Morse theory is used to build a fundamental relationship between caging loops and Morse saddle points.


\section{Theory}\label{sec:theory}
\subsection{Grasping Strategy}
In order to define a caging loop, we have to {consider at least geometric and mechanical aspects.} The geometric considerations include:
\begin{itemize}
\item A caging loop encompasses the target object - any penetration into the target shape is not allowed.
\item A caging loop encloses some part of the target object tightly, i.e., cannot be shortened with respect to a slight perturbance (i.e., {\em stable grasp}) or at least goes around the target object like a great circle enclosing a sphere (i.e., {\em unstable grasp}).
\item A caging loop should roughly match the real robot gripper size.
\end{itemize}
On the other side, the mechanical considerations include
\begin{itemize}
\item The center point of a caging loop should be as close as possible to the center of gravity of the target shape so as to minimize the moment of intertia.
\item A caging loop should be roughly horizontal so that the target object can be taken up steadily.
\end{itemize}

Our strategy is to compute a collection of caging loop candidates in consideration of the above-mentioned geometric principles. For purpose of efficient computation, we also invent a set of filtering techniques to reduce the number of loop candidates as far as possible.

\subsection{Mathematical Formulation}
Imagine the scenario of a caging grasp where the fingers of the gripper stretch to two opposite directions, roughly forming a loop; See Fig.~\ref{fig:teaser} and Fig.~\ref{fig:topologyAndGeometry}.
In the following, we shall formally characterize in which space we extract caging loops and systematically establish properties of caging loops.

The surface $S$ of the target object, typically represented as a watertight mesh, divides the whole $\mathbb{R}^3$ space into interior parts and exterior parts, where only the visible free space (the outmost surface exterior space) is helpful to determine a real grasp configuration.
Rather than constrain caging loops lying on the target surface $S$, we relax caging loops from $S$ to the shape embedding space.

\begin{defn}The visible free space separated by the target surface $S$ is called the {\em grasping space}.
\end{defn}

Generally speaking, a stable grasp is desired, i.e., the caging loop encloses some part of the target object tightly and  cannot be shortened even with a slight perturbance. In some rare cases, however, an unstable grasp like a great circle enclosing a sphere is also acceptable. Both cases imply that there is a fundamental relationship between caging loops and locally shortest loops in the grasping space.
\begin{defn}\label{defn:grasping_loop}Suppose the target object $S$ defines a grasping space $\mathbf{G}$. A closed curve $L\in \mathbf{G}$ is called a {\em caging loop candidate} if and only if $L$ is {\em locally shortest everywhere}, i.e., for any point $p\in L$, any sufficiently short segment of $L$ around $p$ cannot be shortened any more. All such loop candidates constitute a {\em caging loop space}, denoted by $\mathbf{L}$.
\end{defn}

\begin{property}\label{property:three}Each caging loop $L\in{\mathbf{L}}$ touches the target surface at three or more points.
\end{property}

 {\em Proof.}
Without loss of generality, we assume that $L$ touches the target surface $S$ at only one point $p$. Then the open curve ${{L}}\backslash p$ lies in the grasping space but doesn't touch $S$. Considering that a locally shortest curve in $\mathbb{R}^3$ must be a straight line segment, ${{L}}\backslash p$  cannot include a bending point. Furthermore, $p$ is not only the start point of the straight line segment ${{L}}\backslash p$ but also its endpoint. Therefore, $L$ degenerates into a single point under the above assumption, which contradicts to the given condition that $L$ is a caging loop. Similarly, it can be shown that the case of two touching points is impossible.

We can further show that each caging loop consists of an alternative sequence of straight line segments in the grasping space and geodesics on the target surface.
\begin{property}Let ${S}$ be the target surface. Each caging loop $L$ consists of an alternative sequence of geodesic paths lying on ${S}$ and straight line segments in the grasping space $\mathbf{G}$, where a geodesic segment may degenerate into a single point.
\end{property}

In fact, the loop space $\mathbf{L}$ includes all geodesic loops constrained on the surface $S$ and thus cannot be empty, which can be easily verified from the Lusternick-Schnirelmann theorem~\cite{Lusternik1934M}.
\begin{theorem} ${\mathbf{L}}$ is non-empty.
\end{theorem}

However, it is difficult to directly extract a caging loop without any further hint. Therefore, we consider a type of relaxed caging loops.

\begin{defn}\label{defn:relaxed_loop}Suppose the target object $S$ defines a grasping space $\mathbf{G}$. Let $p$ be a point in $\mathbf{G}$.  A closed curve $L\in \mathbf{G}$ is called a {\em $p$-based caging loop candidate} if and only if $L$ is { locally shortest everywhere} except at $p$. When $p$ is taken over all points in $\mathbf{G}$, all such loop candidates constitute a different {caging loop space}, denoted by $\widetilde{\mathbf{L}}$.
\end{defn}

We immediately have the following property.
\begin{property}$\widetilde{\mathbf{L}}$ is a superset of ${\mathbf{L}}$.
\end{property}

{\bf Remark:} Each loop $L\in\widetilde{\mathbf{L}}$ carries a base point $p$. If we eliminate those loops that are not locally shortest at the corresponding base point, then $\widetilde{\mathbf{L}}$ becomes ${\mathbf{L}}$. Therefore, the above property implies that we can select the desirable caging loop from $\widetilde{\mathbf{L}}$.

\section{Methodology}\label{sec:methodology}
\subsection{Computing Loop Candidates}
\begin{figure}[ht]
  \centering
  \includegraphics[width=0.85\columnwidth]{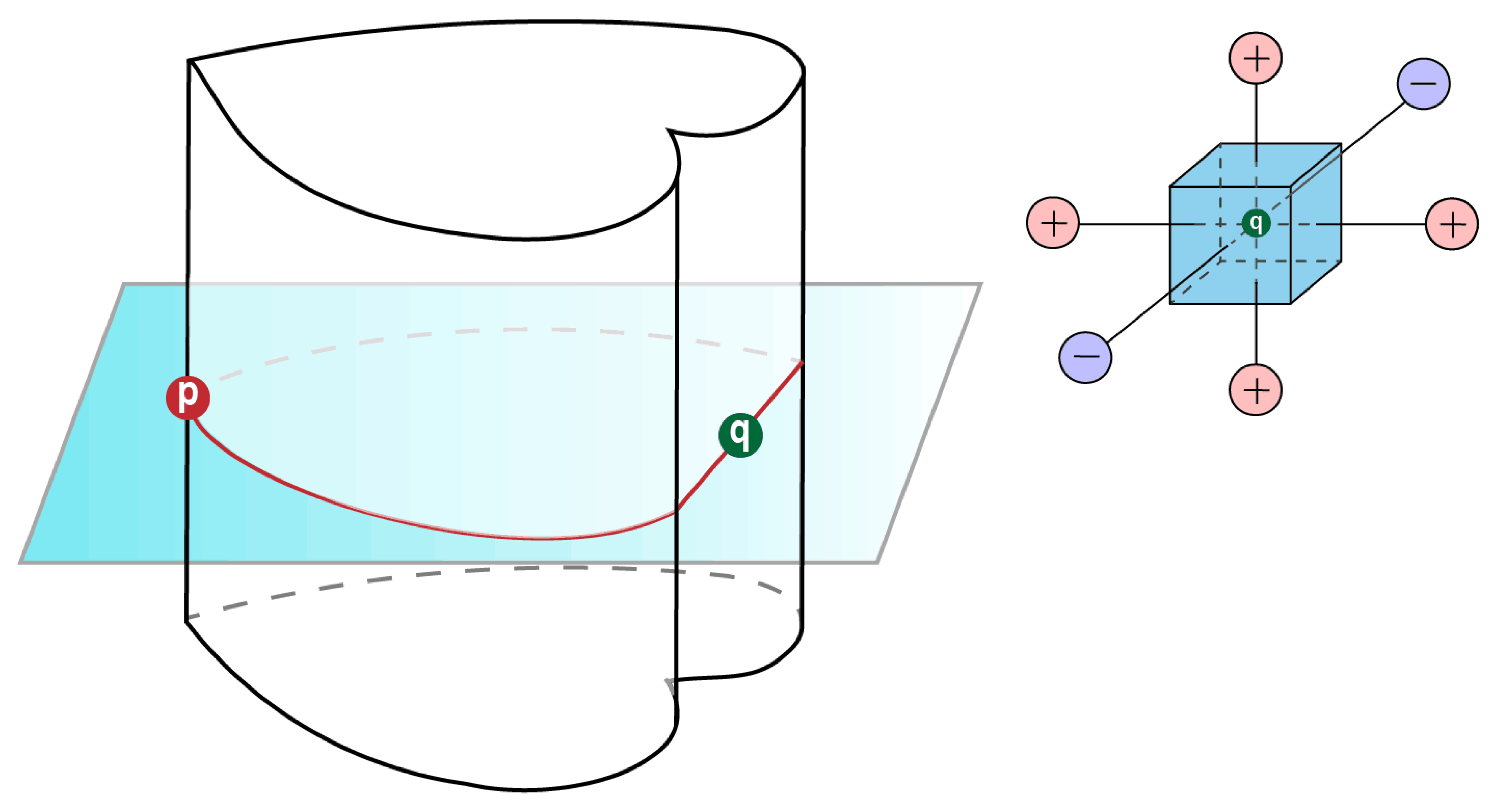}
  \caption{\label{fig:morse_saddle}An example of Morse-Smale saddle point of the $p$-based distance field restricted in a grasping space $\mathbf{G}$.
   }
   \vspace{-2mm}
\end{figure}

Let $p$ be a point in the grasping space $\mathbf{G}$. For each point $x$ in $\mathbf{G}$, we use $\mathbf{D}_p(x): G \rightarrow \mathbb{R}$ to denote the length of the shortest path connecting $p$ and $x$. $\mathbf{D}_p$ is called the distance field rooted at $p$. Note that the distance is measured in $\mathbf{G}$ rather than on the target surface.

Suppose that $L_p$ is a $p$-based loop in the caging loop space $\widetilde{\mathbf{L}}$.
It is easy to know that there is a point $q\in L_p$ such that $q$ divides $L_p$ into two equal-length parts. Obviously, both the two sub-curves are locally shortest paths in the grasping space $\mathbf{G}$.
In the following, we shall reveal the fact that there is a fundamental relationship between Morse theory and caging loops.

\begin{theorem}\label{thm:saddle}Suppose that the target object defines a grasping space $\mathbf{G}$. Let $\mathbf{D}_p$ be the distance field rooted at $p\in\mathbf{G}$. Each Morse saddle or maximal point of $\mathbf{D}_p$ is able to define a $p$-based caging loop.
\end{theorem}

 {\em Proof.}
 Let $q$ be an Morse saddle (or maximal) point of $\mathbf{D}_p$. Since there exists a pair of shortest paths $\Pi_1, \Pi_2$ that go along opposite directions  at $q$. By combining $\Pi_1, \Pi_2$, we get a  $p$-based caging loop.

Although the above discussion is assumed in the continuous setting, Morse-Smale theory is also well defined in the discrete setting; see more details in~\cite{Ni-2004FairMF}. We can inherit the spirit in~\cite{Ni-2004FairMF} and distinguish Morse saddle points, minimal points and maximal points by considering the relative magnitude at a voxel and its neighboring voxels. As Fig.~\ref{fig:morse_saddle} shows, we label $p$'s neighbor with a ``+'' if the neighbor has a higher value and a ``-'' otherwise. It is easy to know that there are at most $2^6$ possible configurations. Fig.~\ref{fig:morse_saddle} gives a typical situation of Morse saddle point, where two opposite neighbors are labeled with ``-'' while the other four neighbors are labeled with ``+''. Similarly, a voxel is classified as a maximal point if all 6 neighbors are labeled with ``-''.

Based on Theorem~\ref{thm:saddle}, it is natural to devise a na\"{i}ve algorithm (see Algorithm~\ref{alg:naive}) to build the $p$-based caging loop space $\widetilde{\mathbf{L}}$.
\begin{algorithm}[htb]
\caption{A na\"{i}ve algorithm  for computing $\widetilde{\mathbf{L}}$\label{alg:naive}}
Initialize $\widetilde{\mathbf{L}}$ to be empty.\\
Compute a sample set $P$ in the grasping space $\mathbf{G}$.\\
\For {each point $p\in P\subset\mathbf{G}$}
{
Compute the distance field $\mathbf{D}_p$.\\
\For {each Morse saddle or maximal point $q$ of $\mathbf{D}_p$}
{
Trace a $p$-based caging loop.\\
Add it into $\widetilde{\mathbf{L}}$.
}
}
\end{algorithm}
%

\subsection{Filtering Rules}
However, $\widetilde{\mathbf{L}}$ is very large generally. We need to invent a handful of filtering rules to reduce the computational cost.
First of all, the reduction of the grasping space $\mathbf{G}$ is much helpful to filter out redundant caging loops.
\begin{theorem}\label{thm:convexhull}Let ${H}$ be the convex hull of the target surface $S$. Any caging loop must lie between $S$ and $H$.
\end{theorem}

{Secondly}, Property~\ref{property:three} asserts that the base point $p$ can be constrained on the target surface $S$, which cannot cause missing any useful caging loop. In fact, the location of $p$ can be more restricted; See the following theorem.
\begin{theorem}\label{thm:principal}For a point $p$ on the target surface $S$, if both the principal curvatures are negative, $p$ cannot determine a caging loop.
\end{theorem}

{\em Proof.}
Suppose $L_p$ is a caging loop. The sufficiently short segment of $L_p$ around $p$ can be viewed as the intersection between a normal plane at $p$ and the target surface $S$. Since both the principal curvatures at $p$ are negative, the loop can be shortened by moving $p$ toward $\mathbf{G}$ a little bit, leading to a contradiction.


{Finally}, even if the base point $p$ is given, there is no need to compute the entire distance field  $\mathbf{D}_p$ since an overly long loop is no use for grasping. It is sufficient to limit the sweep process in an appropriate range comparable to the gripper size. In practice, it is reasonable to require that the total stretching length of the robot gripper (twice as long as the gripper finger), denoted by $2h$, should be larger than one half of the length of the caging loop. Therefore, during the computation of the $p$-based distance field, we can terminate the sweep process when the sweep radius amounts to $2h$ since at this moment, any $p$-based caging loop longer than $4h$ has been found.

Taking the above speedup techniques simultaneously into consideration, we give an advanced algorithm for computing caging loops; See Algorithm~\ref{alg:advanced} (the difference from Algorithm~\ref{alg:naive} is underlined).
\begin{algorithm}[htb]
\caption{An advanced algorithm for computing $\widetilde{\mathbf{L}}$\label{alg:advanced}}
Initialize $\widetilde{\mathbf{L}}$ to be empty.\\
Compute the grasping space $\mathbf{G}$ {\underline {between the target}} {\underline {surface $S$ and its convex hull; see Theorem~\ref{thm:convexhull}}}. \\
\For {\underline{each sample point $p\in S$ with one of the principal } \underline{curvatures being positive (see Theorem~\ref{thm:principal})}}{
Compute  $\mathbf{D}_p$ {\underline{with a sweep radius of $2h$}}.\\
\For {each Morse-saddle point $q$ of $\mathbf{D}_p$}
{
Trace a $p$-based caging loop.\\
Add it into $\widetilde{\mathbf{L}}$.
}
}
\end{algorithm}
%

\section{Implementation}\label{sec:implementation}
\begin{figure*}[ht]
  \centering
        \includegraphics[width=1.0\linewidth]{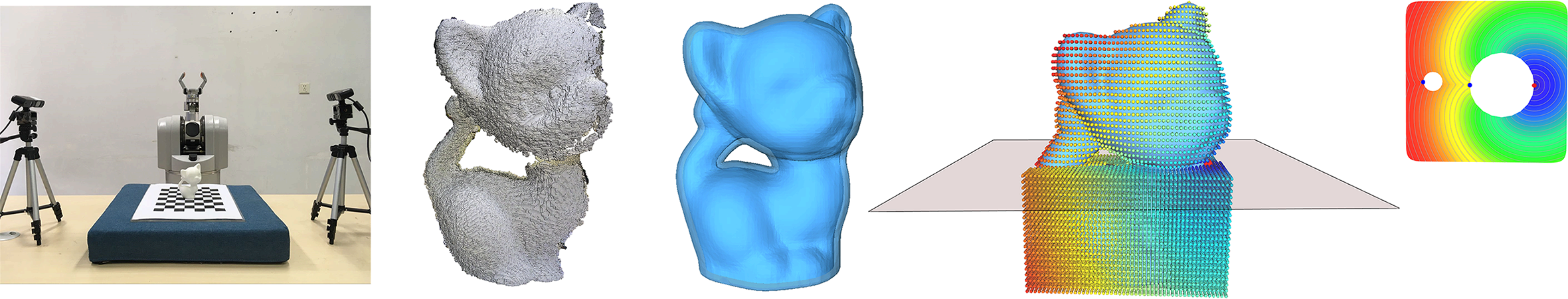}
          \parbox[t]{1.0\linewidth}{\relax
           \hspace{17 mm} (a) \hspace{33 mm} (b) \hspace{23 mm} (c) \hspace{60 mm} (d)}\\
           \parbox[t]{1.0\linewidth}{\relax }\\
        \includegraphics[width=1.0\linewidth]{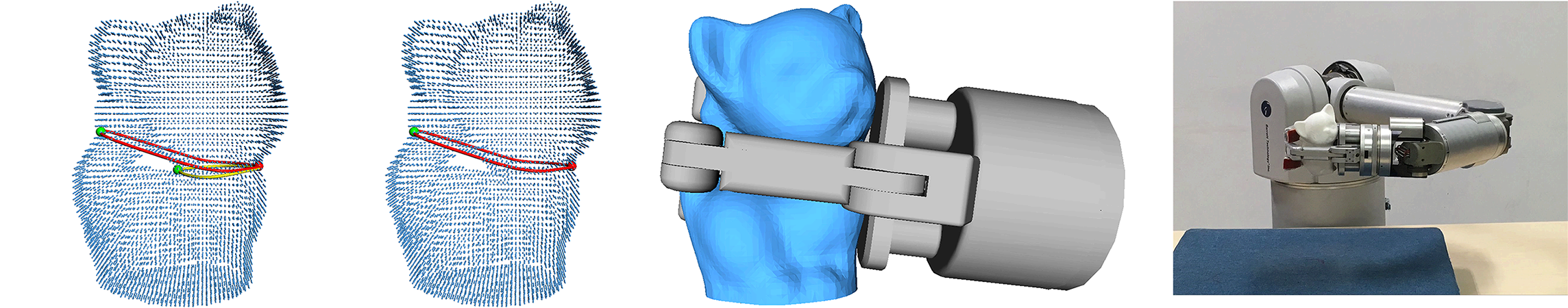}
         \parbox[t]{1.0\linewidth}{\relax
           \hspace{17 mm} (e) \hspace{33 mm} (f) \hspace{23 mm} (g) \hspace{60 mm} (h)}\\
        \caption{
         \label{fig:overview}
            An overview of our caging loop based grasping system.
            (a) Our system setup, composed of one robotic arm and two depth cameras.
            (b) The incomplete point cloud scanned by the two depth cameras.
            (c) The $r$-offset surface of the reconstructed target object that defines the grasping space.
            (d) A $p$-based distance field and two Morse saddle points (blue).
            (e) Two caging loop candidates induced by the two Morse saddle points.
            (f) The yellow loop is filtered since it is far from being locally shortest at the base point (red).
            (g) A simulation of grasping.
            (h) Real grasping conducted by our system.
        }
   \vspace{-4mm}
\end{figure*}

In a real grasping scenario, the gripper thickness cannot be negligible - a gripper cannot stretch into small topological holes or gaps. A commonly used technique is to filter out those infeasible caging loops by checking interference. Rather than leave it to ex post interference analysis, in this paper, we take each gripper finger as a skeleton curve equipped with a sweep radius $r$. In implementation, we offset the target surface outward in a distance of $r$ and require any caging loop to be lying in the grasping space $\mathbf{G}_r$ separated by the $r$-offset surface $S_r$. Fig.~\ref{fig:overview} shows an implementation details of our approach.

\subsection{Grasping Space $\mathbf{G}_r$}\label{subsec:GraspingSpace}
Given a scanned point cloud $\{(x_i,\mathbf{n}_i)\}$ of the target shape, we shall adapt the radial basis
function (RBF) technique ~\cite{Carr-2001ReconstructionAR} to represent the $r$-offset surface $S_r$, which is central to define the grasping space $\mathbf{G}_r$. The general RBF with regard to $\{(x_i,\mathbf{n}_i)\}$ is defined as follows:
$$
f(x) = \sum_{i}^n w_i\phi(\|x-x_i\|)+P(x),
$$
where $\phi(t) = t$ is the basis function used in our experiments, and the weighting coefficients $\{w_i\}$ and the low-degree (typically linear) polynomial $P(x)$ is undetermined.
Taking $x_j$ into the RBF, we have
$$
\sum_{i}^n w_i\phi(\|x_j-x_i\|)+(1,x_j)^\text{T}\mathbf{c}=f(x_j) = 0.
$$
Furthermore, considering that the point $x_j+r\mathbf{n}_j$ that lies on the $r$-offset surface $S_r$,, we have
$$
\sum_{i}^n w_i\phi(\|x_j+r\mathbf{n}_j-x_i\|)+(1,x_j)^\text{T}\mathbf{c}=f(x_j+r\mathbf{n}_j) = r,
$$
where $\mathbf{c}=(c_0,c_1,c_2,c_3)$ is unknown.
At the same time, in the RBF based approach, the four side conditions are
$$
\sum_{i}^n w_i = \sum_{i}^n w_i x_i = \sum_{i}^n w_i y_i = \sum_{i}^n w_i z_i = 0.
$$
The above formulation can be finally transformed into a linear system from which we can immediately compute $\{w_i\}$ and $P(x)$. To this end, we find an implicit surface $f(x)=r$ to represent the $r$-offset surface $S_r$. If the shape embedding space is discretized into voxels, it is very easy to identify outside voxels that meet $f(x)\geq r$, which can be viewed as a discrete representation of the grasping space $\mathbf{G}_r$.
\subsection{Gripper Configuration}\label{subsec:GripperConfiguration}
\begin{figure}[ht]
  \centering
  \includegraphics[width=0.95\columnwidth]{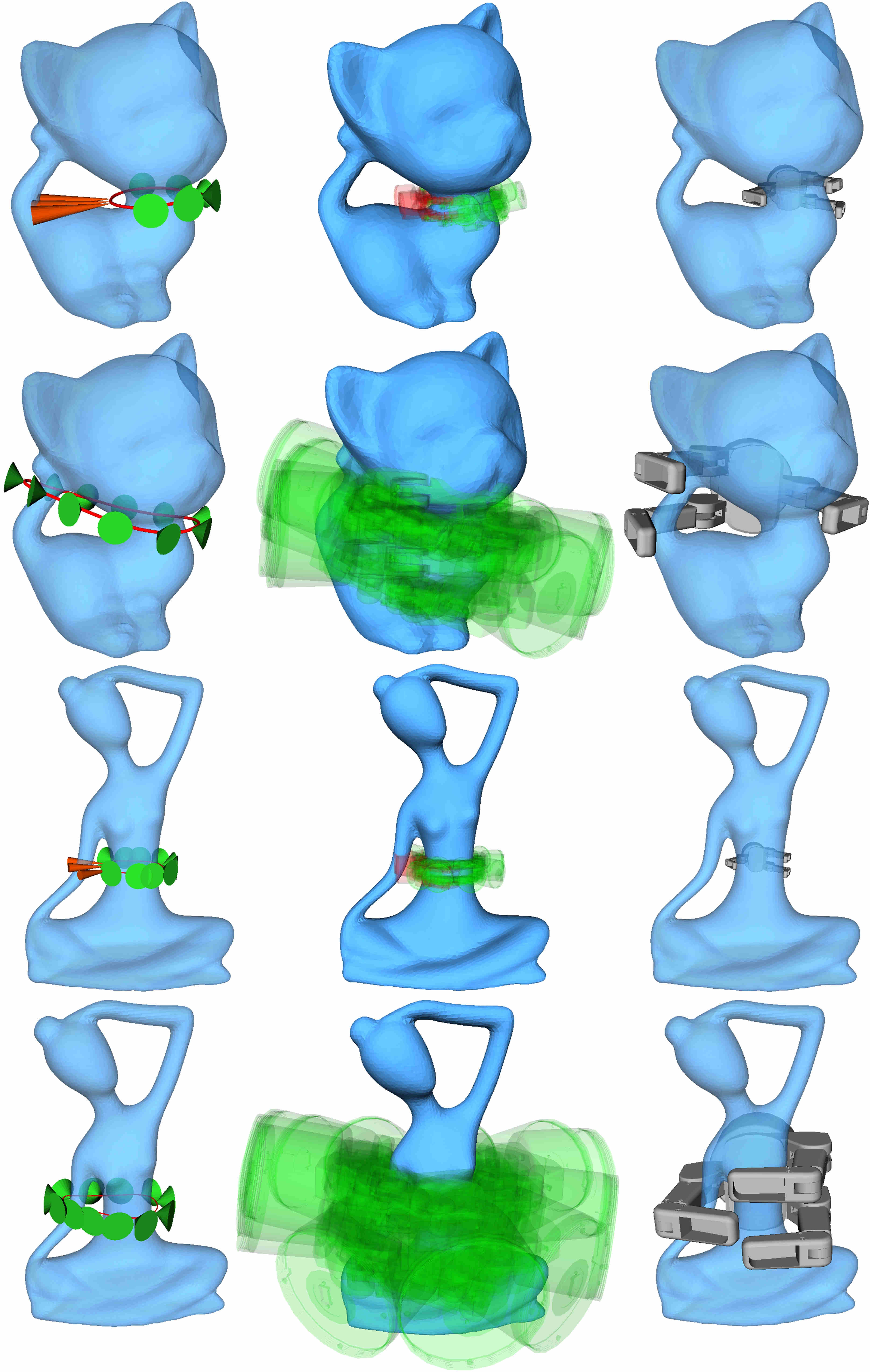}
  \caption{\label{fig:gripperConfiguration}
    Inferring the gripper configuration based on a caging loop. In order to define a valid caging configuration,
    we check global interference in the simulation environment of OpenRAVE. Row 1: Large-size Kitten; Row 2: Small-size Kitten; Row 3: Large-size Yoga; Row 4: Small-size Yoga.
   }
   \vspace{-2mm}
\end{figure}

Upon obtaining a desirable caging loop $L$, we need to determine the origin position $o$ of the gripper, as well as an orthogonal frame to set the gripper orientation, which facilitates a real grasp.
Suppose that $L$ is represented by a point sequence $\{p_1,p_2,\cdots,p_n\}$.
Imagine that there is an inward cone rooted at $p\in\{p_1,p_2,\cdots,p_n\}$ and the center line of the cone coincides with the normal vector at $p$; See Fig.~\ref{fig:gripperConfiguration}. (If $p$ is not located on $S_r$, the normal vector is discussed later.) We further define $\theta_p$ to be the maximum open angle under the condition that no penetration occurs between the inward cone and the target shape. The origin point $o$ is then selected from $\{p_1,p_2,\cdots,p_n\}$ so as to maximize the opening angle.
Let $c=\frac{p_1+p_2+\cdots+p_n}{n}$ be the center point of $L$.
We then define the first direction $Dir_1$ as follows:
$$
Dir_1 = \frac{o-c}{\|o-c\|},
$$
which roughly means the forward direction of the gripper. Considering that the loop $L$ is roughly a planar curve, we can fit $L$ using a plane
$
\mathbf{n}\cdot x = b,
$
where $\mathbf{n}$ is a unit vector. Finally, we define $Dir_2$ as follows:
$$
Dir_2 = Dir_1 \times \mathbf{n}.
$$
If the triple $(o, Dir_1, Dir_2)$ is able to define a valid grasping configuration (no global interference happens), we use it to guide the orientation of the gripper and launch a real grasp. In our experiments, the gripper will spread its fingers and move to the point $o$ first. It then wraps the target object tightly with the hint of $Dir_1$ and $Dir_2$.

In Fig.~\ref{fig:gripperConfiguration}, we show some examples on how to find a valid grasping configuration assuming that a desirable caging loop has been found. The key step is to check global interference in the simulation environment of OpenRAVE. It can be observed that our approach can report different gasping configurations on the same model with different sizes.

%

{\bf Remark: }If $p$ is not located on $S_r$, $Dir_1$ is given by
$\mathbf{t}_p\times \mathbf{n}$, where $\mathbf{t}_p$ is the tangent direction of the loop $L$ at $p$ and $\mathbf{n}$ is the normal to the fitting plane of $L$.

\section{Experimental Results}\label{sec:experimental}
We conducted both simulation and mechanical experiments to validate our approach.
{In this section, we first test the effectiveness of our algorithm on a variety of complex objects. We then show that our algorithm can be applied to 3D shapes with various levels of noise and geometric features. After that, we demonstrate the superior caging ability of our approach on real grasping scenarios (the digital models of the target objects are unknown in advance).
Finally, we give the timing statistics of the main computational steps.}
\subsection{Test on High-genus Models in Various Sizes}
\begin{figure}[ht]
  \centering
  \includegraphics[width=0.95\columnwidth]{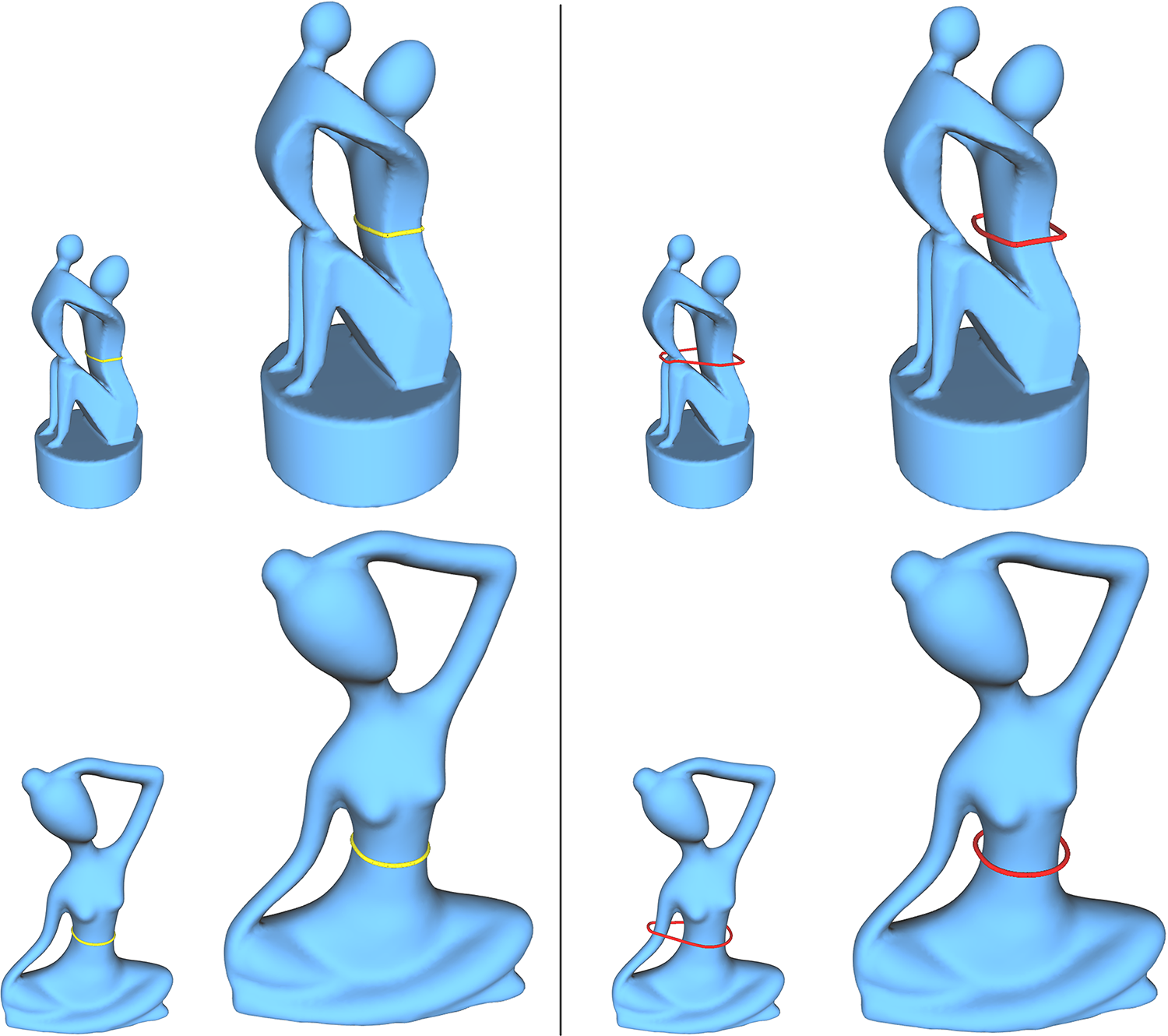}\\
    \parbox[t]{1.0\columnwidth}{\relax
           \hspace{17 mm} (a)
           \hspace{31 mm} (b)
           }
  \caption{\label{fig:adaptive} Caging loops generated on the Fertility and Yoga models with various sizes.
  (a) Caging loops (yellow) produced by the method in~\cite{Zarubin-2013CagingCO}.
  (b) Caging loops (red) computed by our method.
   }
   \vspace{-2mm}
\end{figure}
\begin{figure*}[ht]
  \centering
  \includegraphics[width=0.95\linewidth]{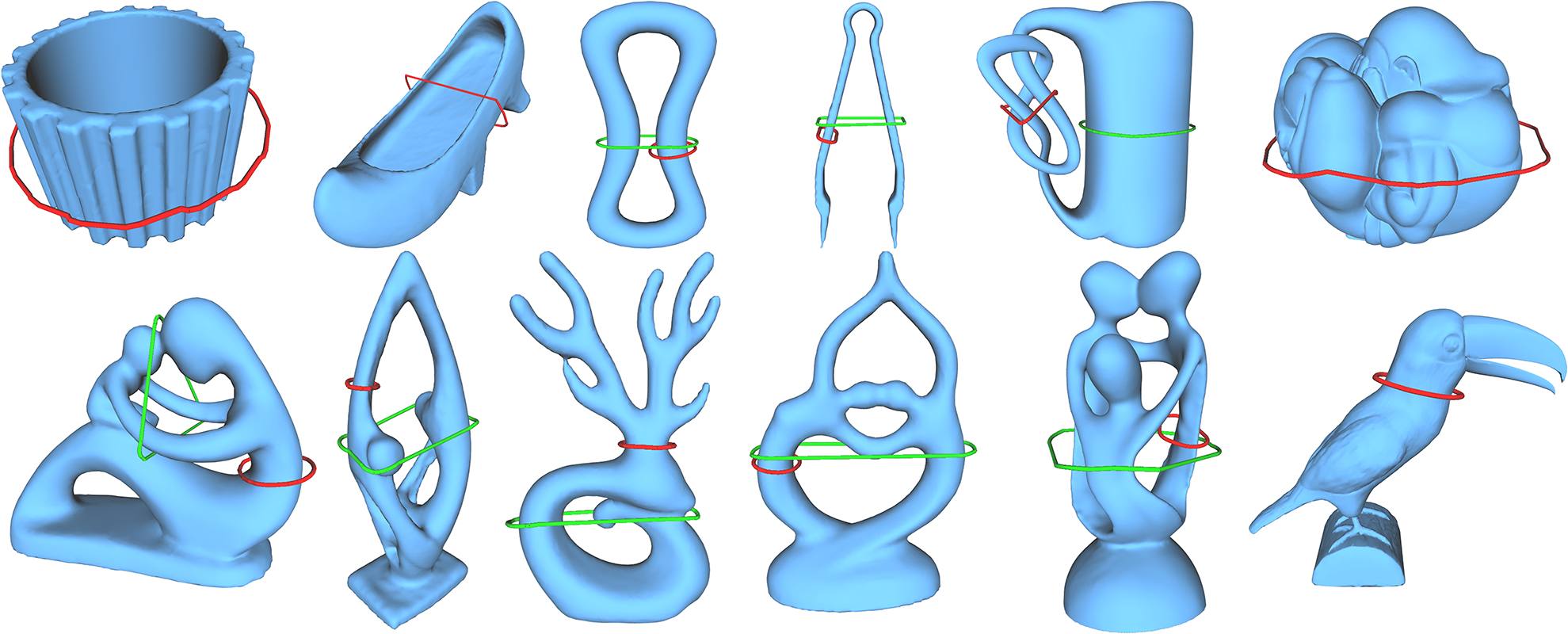}
  \caption{\label{fig:MoreGraspingLoops}
    {A gallery of caging loops computed for a variety of 3D models.
    For models with multiple topological handles, we show two caging loops.
    The green loop is computed for a relatively large gripper enclosing multiple handles.
    The red loop, on the other hand, is for a small gripper to grasp a handle.}
   }
   \vspace{-2mm}
\end{figure*}
There are a number of research works on caging a 3D shape, which
largely fall into one of the following three categories:
topology guided~\cite{Pokorny-2013GraspingOW}, geometry guided~\cite{Zarubin-2013CagingCO} and topology \& geometry guided~\cite{Kwok-2016RopeCA,Varava-2016CagingGO}.
Existing approaches, whether topology guided or geometry guided, consider only those loops constrained on the target surface. Therefore, it is hard for them to deal with small topological holes or small gaps.
As shown by the comparison in Fig.~\ref{fig:adaptive}, our algorithm handles 3D shapes with small topological holes and is aware of the relative size between the gripper and the shape.
More caging loop examples can be found in Fig.~\ref{fig:MoreGraspingLoops}.
\subsection{Test on Models with Various Levels of Noise and Geometric Feature}
\begin{figure}[ht]
  \centering
  \includegraphics[width=\columnwidth]{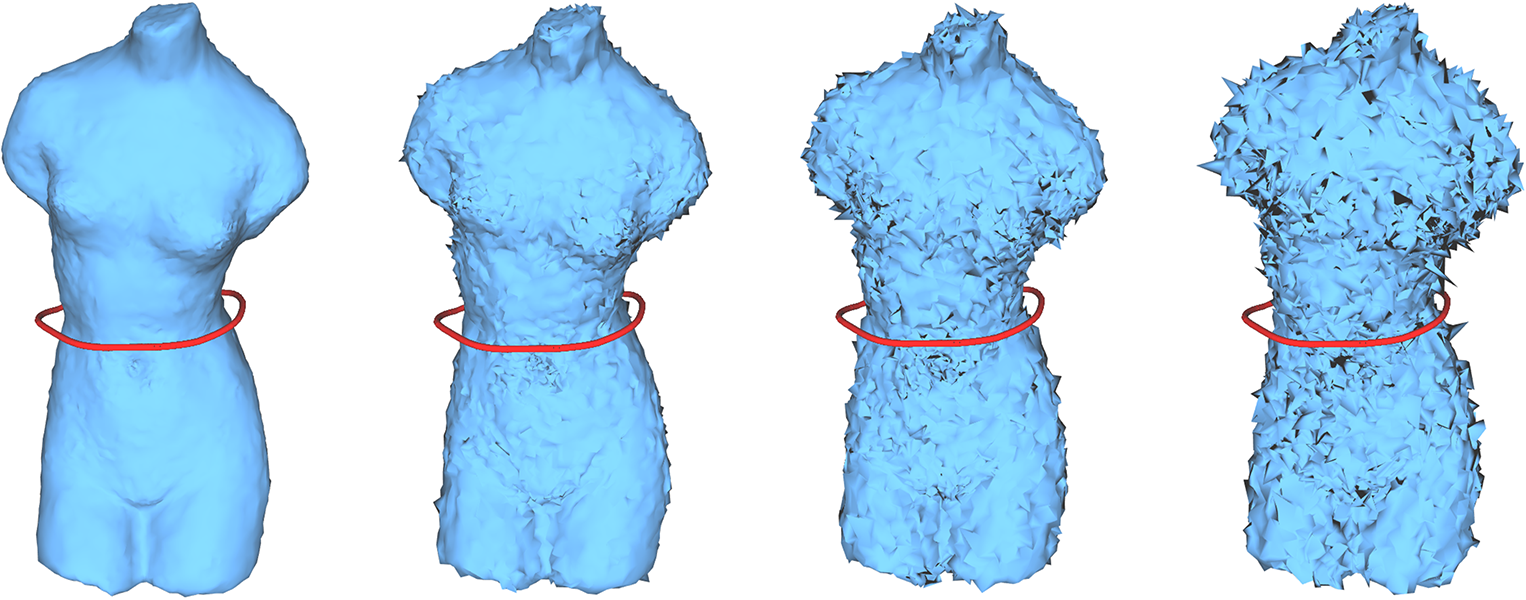}\\
  \includegraphics[width=\columnwidth]{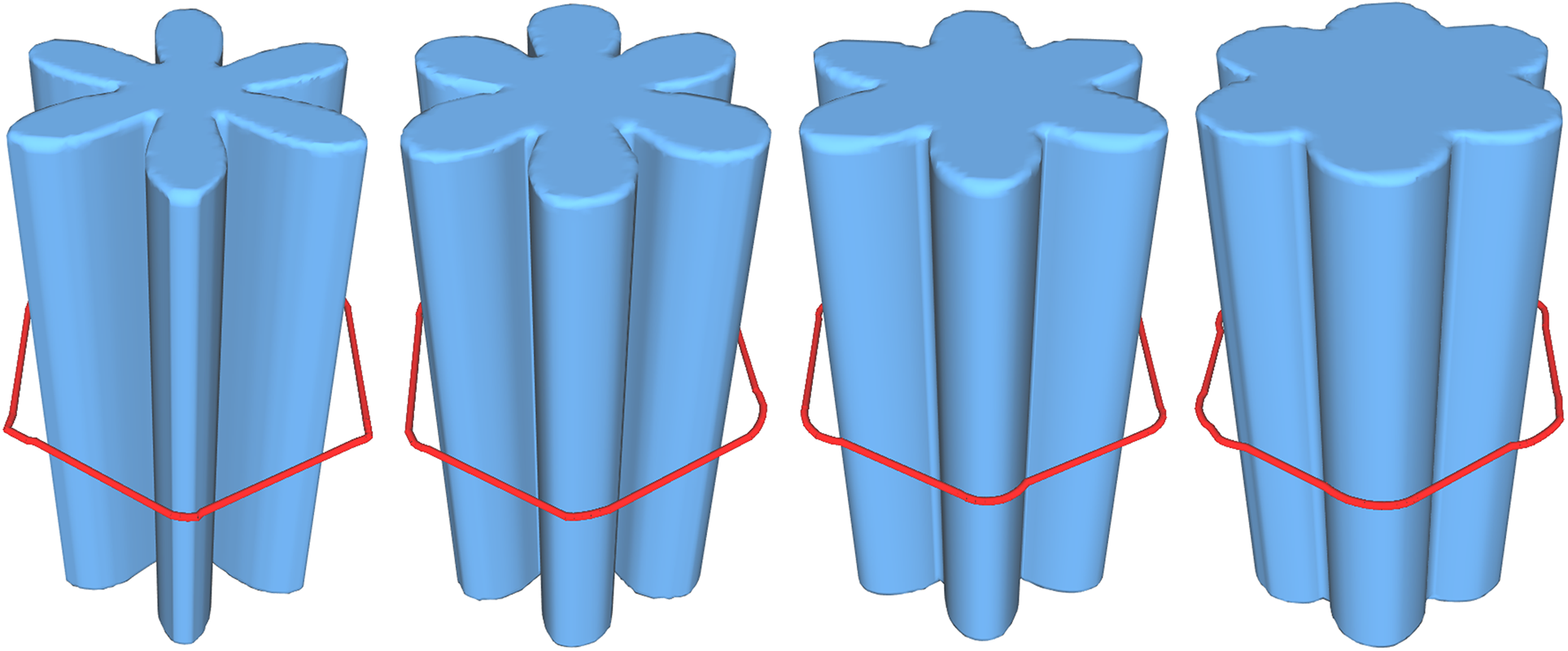}
  \caption{\label{fig:venusbodynoise}
    {Top row: Caging loops on the Venus model with varying levels of Gaussian noise (relative to the whole model size): 0.01, 0.03, 0.05, 0.07.
     Bottom row: Caging loops computed by our method on the Pillar models are insensitive to varying levels of geometric details.}
   }
   \vspace{-2mm}
\end{figure}
In real grasping scenarios, the target object is often scanned into a point cloud with noise. Therefore, a key criteria to evaluate a caging algorithm is whether it is robust to geometric noise or variations. In Fig.~\ref{fig:venusbodynoise}, the top row shows a group of caging loops on the Venus models with various levels of noise, while bottom row shows a group of caging loops on the Pillar models with various levels of geometric details. Both of them exhibit the robustness of our algorithm against geometric noise and details.
\subsection{Test on Real Objects}
\begin{figure}[t]
  \centering
  \includegraphics[width=1.0\columnwidth]{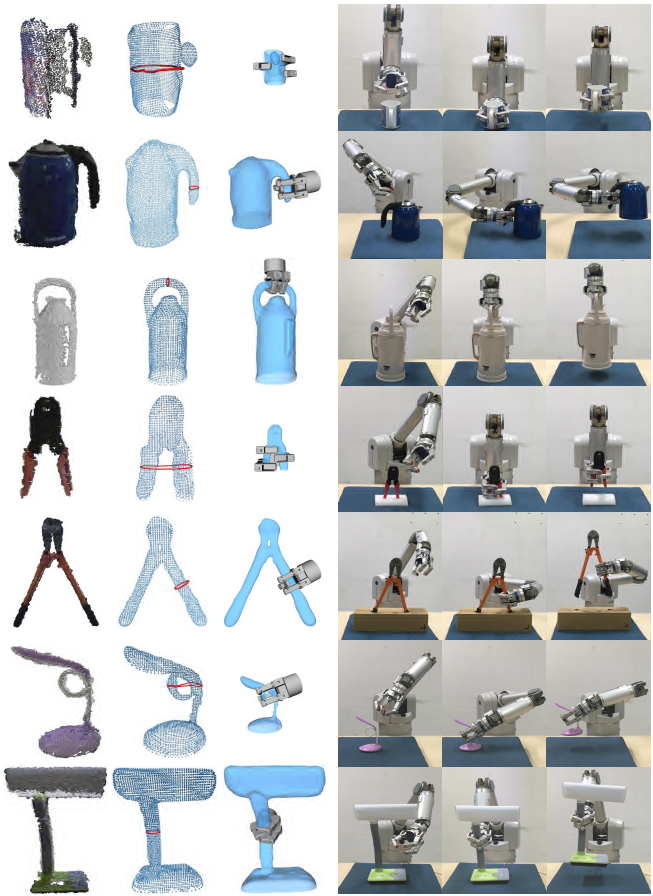}\\
  \caption{\label{fig:realDataResult}
   Grasping household objects by our system.
   From left to right: Scanned point clouds, offset surfaces and caging loops,
   simulation results, progressive demonstration of real grasping.
   }
  \vspace{-2mm}
\end{figure}
In order to validate our approach on real data, we build a platform with a 7-DoF Barrett WAM arm, a Barrett BH8-282 three-finger gripper and two Xtion Pro depth cameras. For each object shown in Fig.~\ref{fig:realDataResult}, we keep the depth cameras unchanged while rotating the target shape repeatedly for $10$ times. In this way, we recorded of the grasp success rates. A grasp is regarded to be successful if the target shape does not escape from the gripper during a large-scale movement that is about $10$cm off the ground. 
We report the success rate of grasping in Table~\ref{tab:GraspSuccessfulRate}.
It shows that our approach can synthesize reliable caging grasps, as compared to traditional approaches whose success rate is generally about $85\%$.
\begin{table}[ht]
\centering
\caption{Grasping success rate.}
\resizebox{0.7\columnwidth}{!}
{
\begin{tabular}{|r|c|}
\hline
Target Object                                      & Success rate   \\ \hline
Mug             (Fig.~\ref{fig:realDataResult})  &100\%           \\ \hline
kettle          (Fig.~\ref{fig:realDataResult})  &100\%            \\ \hline
Big Thermos     (Fig.~\ref{fig:realDataResult})  &100\%            \\ \hline
Pliers          (Fig.~\ref{fig:realDataResult})  &90\%           \\ \hline
Big Pliers      (Fig.~\ref{fig:realDataResult})  &90\%                \\ \hline
Small Lamp      (Fig.~\ref{fig:realDataResult})  &90\%           \\ \hline
Big Lamp        (Fig.~\ref{fig:realDataResult})  &100\%           \\ \hline
\end{tabular}
}
\label{tab:GraspSuccessfulRate}
\end{table}

\subsection{Performance}
In order to accomplish a grasping task, we have to perform a sequence of shape analysis operations and then send a grasp instruction to the robot. Recall that we define a grasping space and search the best caging loop in that space.
In our implementation, we discreticize the bounding box enclosing the target shape into $50 \times 50 \times 50$ voxels and label each voxel between the $r$-offset surface $S_r$ and its convex hull $H_r$ with ``1''. After that, we extract a set of $500$ uniformly distributed sample points from $S_r$ and eliminate those sample points that do not help determine a caging loop at all (see Theorem~\ref{thm:principal}). Finally, the loop candidate pool is generated based on Morse theory. Among these steps, the most time-consuming steps include
the grasping space computation, a distance field generation for a collection of base points and topological analysis based on Morse theory.
The mesh models shown in this paper are discretized into 2K vertices, and the average computation time for each model is about 1.5 seconds. It can be seen that our algorithm runs very fast at this level of resolution. Therefore, if we simultaneously execute the computation task and the move of gripper, it does not introduce a noticeable delay.

\section{Conclusion and Discussion}
In this work, we propose to synthesize feasible caging grasps in the shape embedding space of the target object.
Our caging loops are able to encompass multiple small topological handles and concave regions,
which are relatively too small to be grasped,
through decoupling their computation from surface geometry of the target object.
This also facilitates grasp synthesis for unknown objects which are acquired and reconstructed on-the-fly.
Extensive experimental results exhibit that our approach can deal with real objects with complex surface geometry and topology, being aware of the relative size between objects and gripper.

Our current solution has several limitations.
First, the method used for measuring the physical feasibility is merely a preliminary solution
which can definitely be replaced by other alternatives. Our core method for caging loop computation,
however, ensures the candidate loops are mostly feasible with respect to the gripper size.
Second, the caging loops computed by our method mainly reflect the geometric aspect of graspability
and do not account for the high level information of semantics or functionality.
For example, an object can be grasped in different ways for different purposes.
Synthesizing function related grasps is an interesting venue for future study.

\bibliographystyle{IEEEtran} 
\bibliography{reference}
\end{document}